\documentclass[final]{cvpr}

\usepackage{times}
\usepackage{epsfig}
\usepackage{graphicx}
\usepackage{amsmath}
\usepackage{amssymb}
\usepackage{booktabs}
\usepackage{subfigure}
\usepackage[noend]{algpseudocode}
\usepackage{algorithmicx,algorithm}

\usepackage[pagebackref=true,breaklinks=true,colorlinks,bookmarks=false]{hyperref}

\newcommand{\x}{\boldsymbol{x}}

\def\cvprPaperID{6434} 
\def\confYear{CVPR 2021}

\begin{document}

\title{MetaSAug: Meta Semantic Augmentation for Long-Tailed Visual Recognition}

\author{
	Shuang Li\textsuperscript{1} \space\space\space
	Kaixiong Gong\textsuperscript{1} \space
	Chi Harold Liu\textsuperscript{1}\footnotemark[2] \space\space\space
	Yulin Wang\textsuperscript{2} \space
	Feng Qiao\textsuperscript{3} \space
	Xinjing Cheng\textsuperscript{3} \vspace{.3em}\\
	\textsuperscript{1}Beijing Institute of Technology \quad \textsuperscript{2}Tsinghua University\quad \textsuperscript{3}Inceptio Tech. \\
	\vspace{-.3em}
	{\tt\small \{shuangli, kxgong\}@bit.edu.cn} \space {\tt\small liuchi02@gmail.com} \space {\tt\small wang-yl19@mails.tsinghua.edu.cn} \\
	  {\tt\small feng.qiao@inceptio.ai} \space {\tt\small cnorbot@gmail.com}
	\vspace{-.5em}
}

\maketitle

\pagestyle{empty}  
\thispagestyle{empty} 

\renewcommand*{\thefootnote}{\fnsymbol{footnote}}
\setcounter{footnote}{2}
\footnotetext{C. Liu is the corresponding author.}
\begin{abstract}
   Real-world training data usually exhibits long-tailed distribution, where several majority classes have a significantly larger number of samples than the remaining minority classes. This imbalance degrades the performance of typical supervised learning algorithms designed for balanced training sets. In this paper, we address this issue by augmenting minority classes with a recently proposed implicit semantic data augmentation (ISDA) algorithm \cite{ISDA}, which produces diversified augmented samples by translating deep features along many semantically meaningful directions.
   Importantly, given that ISDA estimates the class-conditional statistics to obtain semantic directions, we find it ineffective to do this on minority classes due to the insufficient training data. To this end, we propose a novel approach to learn transformed semantic directions with meta-learning automatically. In specific, the augmentation strategy during training is dynamically optimized, aiming to minimize the loss on a small balanced validation set, which is approximated via a meta update step. Extensive empirical results on CIFAR-LT-10/100, ImageNet-LT, and iNaturalist 2017/2018 validate the effectiveness of our method.
   
\end{abstract}

\section{Introduction}
\begin{figure}[htbp]
\centering
\includegraphics[width=1\linewidth]{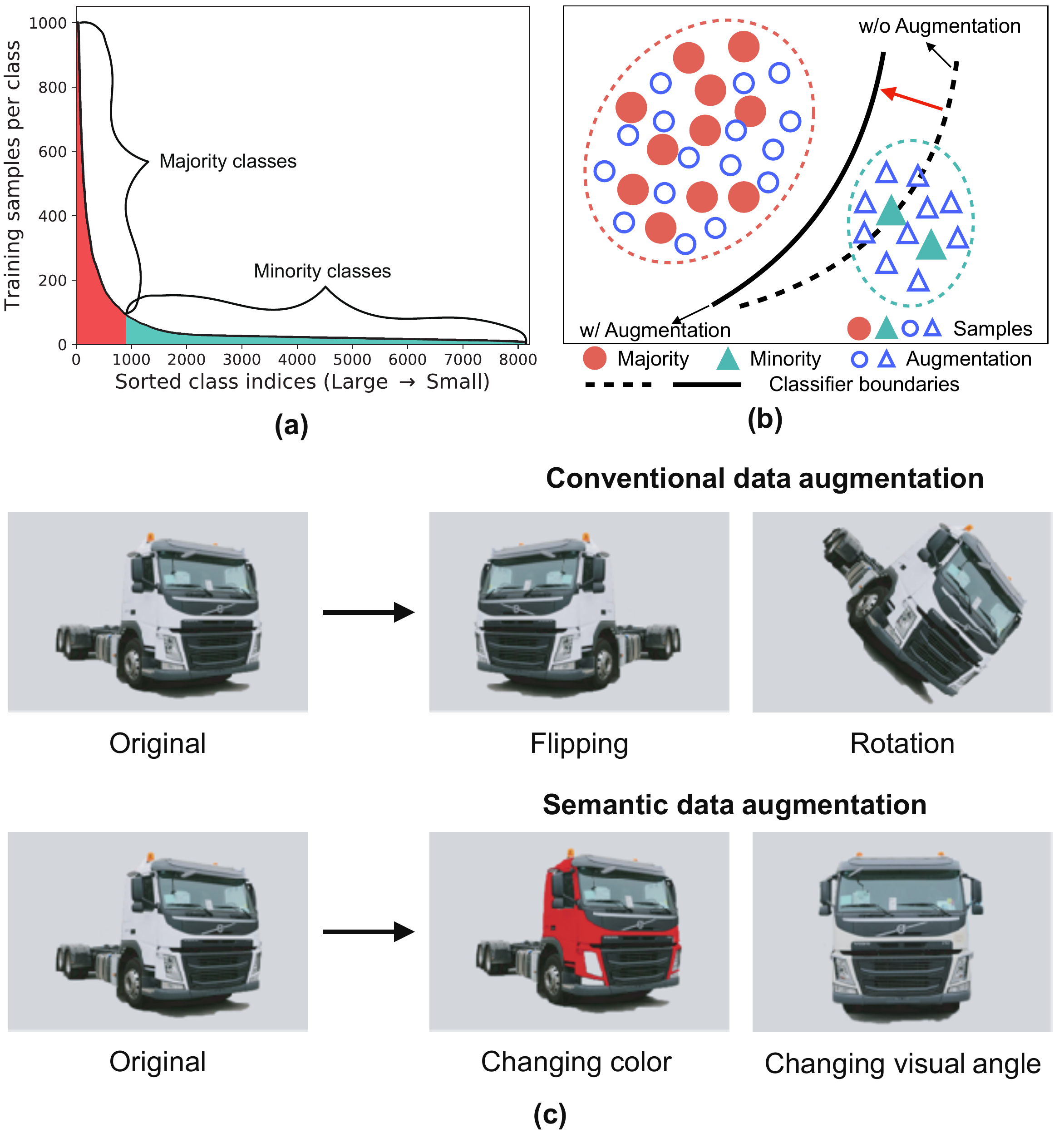}
\caption{\textbf{(a)}: In the data distribution of the real-world dataset iNaturalist 2018, a few majority classes account for the most samples, while the minority classes are under-represented. \textbf{(b)}: Motivation of this work. Facilitating data augmentation for long-tailed problems to ameliorate the classifier performance. \textbf{(c)}: Illustration of traditional data augmentation and semantic data augmentation.}
\label{fig:intro}

\end{figure}
Deep convolutional neural networks (CNNs) have achieved remarkable success in recent years \cite{alexnet,resnet,densenet}. Their state-of-the-art performance is typically demonstrated on the benchmarks such as ImageNet \cite{imagenet2014} and MS COCO \cite{coco}. While these datasets are established by ideally collecting a similar and sufficient number of samples for each class, real-world training data is usually imbalanced, as shown in Fig. \ref{fig:intro}(a). For example, in automatic medical diagnosis, a few common diseases may dominate the training set, with scarce cases for the remaining classes. This long-tailed distribution, unfortunately, degrades the performance of networks severely if using a standard training strategy (e.g., supervised learning with the cross-entropy loss).

To address the issue of data imbalance, a natural solution might be to augment the minority classes for more training samples as shown in Fig. \ref{fig:intro}(b), e.g., by leveraging the data augmentation technique \cite{resnet, densenet, mixup, ratner2017learning}. However, conventional data augmentation techniques like cropping, mirroring and mixup are typically performed on the inputs. As a result, the diversity of augmented samples is inherently limited by the small amount of training data in minority classes. 

Fortunately, this problem can potentially be solved by a recently proposed implicit semantic data augmentation (ISDA) technique \cite{ISDA}. ISDA performs class identity preserving semantic transformation (e.g., changing the color of an object and changing the visual angles) by translating deep features towards certain meaningful semantic directions as shown in Fig. \ref{fig:intro}(c). The deep feature space extracted by CNNs tends to be linearized and has significantly smaller complexity than the pixel space. Therefore, the minority classes will be effectively augmented for more diversity as long as proper semantic directions are found. ISDA estimates class-wise covariance matrices from deep features and sample semantic directions from a Gaussian distribution. Nevertheless, we find that this leads to inferior performance in the long-tailed scenario, since scarce data in minority classes is insufficient to obtain reasonable covariance matrices.

In this paper, we propose a meta semantic augmentation (MetaSAug) approach, aiming to perform effective semantic data augmentation for long-tailed problems via learning more meaningful class-wise covariance. Our major insight here is that \textit{if the appropriate covariance matrices are used for semantic augmentation, the loss on a balanced validation set should be minimized}. At every training iteration, we perform validation on a small balanced validation set, and update the class-wise covariance by minimizing the validation loss. Specifically, we first fulfill the augmentation procedure using current class-wise covariance. Then, we calculate the loss on the validation set with respect to the class-wise covariance. By optimizing the validation loss, we can obtain the updated class-wise covariance that contains rich semantic directions. With it, we train the models on the augmentation set with sufficient semantically augmented samples. In addition, our method can be treated as a plug-in module and be unified with previous methods. We further improve the classification ability of focal loss \cite{lin2017focal} and LDAM loss \cite{cao2019learning} by combining them with MetaSAug.
 
We conduct extensive experiments on several long-tailed datasets, including the artificially long-tailed CIFAR-10/100 \cite{cifar,cui2019class}, ImageNet \cite{imagenet2014,liu2019large}, and the naturally long-tailed dataset inaturalist 2017 and 2018 \cite{inaturalist2017,inaturalist}. The results demonstrate the effectiveness of our method.

\section{Related Work}

In this section, we briefly review the works related to ours.

\textbf{Re-sampling}. Researchers propose to achieve a more balanced data distribution by over-sampling the minority classes \cite{shen2016relay,buda2018systematic,byrd2019effect} or under-sampling the majority classes \cite{he2009learning,japkowicz2002class,buda2018systematic}.
Although being effective, over-sampling might result in over-fitting of minority classes while under-sampling may weaken the feature learning of majority classes due to the absence of valuable instances \cite{multi-label,cao2019learning,chawla2002smote,cui2019class}. Chawla et al. \cite{chawla2002smote} reveal that stronger augmentation for minority classes is beneficial to mitigate over-fitting, which complies with the goal of our method.

\textbf{Re-weighting}. Also termed as cost-sensitive learning, re-weighting aims to assign weights to training samples on either class or instance level. A classic scheme is to re-weight the classes with the weights that are inversely proportional to their frequencies \cite{huang2016learning,wang2017learning}.
Cui et al. \cite{cui2019class} further improve this scheme with proposed effective number. Recently, meta-class-weight \cite{jamal2020rethinking} exploits meta-learning to estimate precise class-wise weights, while Cao et al. \cite{cao2019learning} allocate large margins to tail classes. Apart from above works, Focal Loss \cite{lin2017focal}, L2RW \cite{L2RW} and meta-weight-net \cite{MetaWeightNet} assign weights to examples instance-wisely. Specifically, focal loss assigns weights according to the instance predictions, while L2RW and meta-weight-net allot weights based on the gradient directions. Instead of focusing on designing different weights for classes, our method mainly aims to augment the training set to overcome the imbalance issue.

In addition, for learning better representations, some approaches propose to seprate the training into two stages: representation learning and classifier re-balancing learning \cite{cao2019learning,jamal2020rethinking,cui2018large,kang2019decoupling}. BBN \cite{BBN} further unifies the two stages to form a cumulative learning strategy.

\textbf{Meta-learning and head-to-tail knowledge transfer}. The recent development of meta-learning \cite{MAML,snell2017prototypical,andrychowicz2016learning} inspires researchers to leverage meta-learning to handle class imbalance. A typical series of approaches is to learn the weights for samples with meta-learning \cite{L2RW,jamal2020rethinking,MetaWeightNet}. Another pipeline of methods attempts to transfer the knowledge from head to tail classes. Wang et al. \cite{wang2017learning} adopt a meta learner to regress the network parameters. Liu et al. \cite{liu2019large} exploit a memory bank to transfer the features. Yin et al. \cite{yin2019feature} and Liu et al. \cite{liu2020deep} propose to transfer intra-class variance from head to tail. Different from these works, our method attempt to automatically learn semantic directions for augmenting the minority classes, ameliorating the classifier performance.

\textbf{Data augmentation} is a canonical technique, widely adopted in CNNs for alleviating over-fitting. For example, rotation and horizontal flipping are employed for maintaining the prediction invariance of CNNs \cite{resnet,densenet,simonyan2014very}.  In complementary to the traditional data augmentation, semantic data augmentation that performs semantic altering is also effective for enhancing classifier performance \cite{DAGAN,ISDA}. ISDA \cite{ISDA} performs semantic augmentation with the class-conditional statistics, but cannot estimate reasonable covariance with the scarce data in minority classes. The major difference between ours and ISDA is that MetaSAug utilizes meta-learning to learn proper class-wise covariance for achieving more meaningful augmentation results.

\section{Method}
Consider a training set $D = \{(\boldsymbol{x}_i, y_i)\}_{i=1}^{N}$ with $N$ training samples, where $\boldsymbol{x}_i$ denotes $i$-th training sample and $y_i$ denotes its corresponding label over $C$ class. Let $f$ denote the classifier with parameter $\Theta$ and $\boldsymbol{a}_i$ denote the feature of $i$-th sample extracted by classifier $f$. In the practical applications, the training set $D$ is often imbalanced, resulting in poor performance on the minority classes. Therefore, we aim to perform semantic augmentation for minority classes, ameliorating the learning of classifiers.

\subsection{Implicit Semantic Data Augmentation}
Here, we revisit the implicit semantic data augmentation (ISDA) \cite{ISDA} approach. For semantic augmentation, ISDA statistically estimates the class-wise covariance matrices $\boldsymbol{\Sigma}=\{\boldsymbol{\Sigma}_1, \boldsymbol{\Sigma}_2, ...,\boldsymbol{\Sigma}_C\}$ from deep features at each iteration. Then, ISDA samples transformation directions from the Gaussian distribution $\mathcal{N}(0, \lambda\boldsymbol{\Sigma}_{y_i})$ to augment the deep feature $\boldsymbol{a}_i$, where $\lambda$ is a hyperparameter for tuning the augmentation strength. Naturally, to explore all possible meaningful directions in $\mathcal{N}(0, \lambda\boldsymbol{\Sigma}_{y_i})$, one should sample a tremendous number of  directions. Take a step further, if sampling infinite directions, ISDA derives the upper bound of the cross-entropy loss on all the augmented features:
\begin{align}
    \label{Eq:loss_upper_bound}
    \mathcal{L}_{ISDA} &= \sum_{i=1}^{N} L_{\infty}(f(\boldsymbol{x_i};\Theta),y_i; \boldsymbol{\Sigma}) \\ \notag
    &=\sum_{i=1}^{N} - \log (\frac{e^{z_{i}^{y_i}}}{\sum_{c=1}^{C}e^{z_{i}^{c} + \frac{\lambda}{2} \Delta \boldsymbol{w}_{cy_i}^{\top}\boldsymbol{\Sigma}_{y_i} \Delta \boldsymbol{w}_{cy_i} }}),
\end{align}where $z_{i}^{c}$ is the $c$-th element of the logits output of $\boldsymbol{x}_i$, $\Delta \boldsymbol{w}_{cy_i}= (\boldsymbol{w}_{c}-\boldsymbol{w}_{y_i})$ and $\boldsymbol{w}_{c}$ is the $c$-th column of the weight matrix of last fully connected layer. By optimizing this upper bound $\mathcal{L}_{ISDA}$, ISDA can fulfill the equivalent semantic augmentation procedure efficiently. 

However, the performance of ISDA relies on the covairance matrices estimation. In the long-tailed scenario, we find that ISDA has unsatisfactory performance, since scarce data in minority classes is insufficient to achieve reasonable covariance matrices.

\subsection{Meta Semantic Augmentation}
To address the issue of class imbalance, we propose to augment the minority classes for more training samples. 
As aforementioned, the scarcity of data limits the effectiveness of semantic augmentation. Therefore, we attempt to learn appropriate class-wise covariance matrices for augmenting, leading to better performance on minority classes. The key idea is that \textit{if the appropriate covariance matrices are used for semantic augmentation, the loss on a balanced validation set should be minimized}. In this work, we utilize meta-learning to achieve this goal. 

\textbf{The meta-learning objective}. Generally, by leveraging $L_{\infty}$ in eq. \eqref{Eq:loss_upper_bound}, we can train the classifier and simultaneously fulfill the semantic augmentation procedure. However, in the context of class-imbalanced learning, the majority classes dominate the training set. And from eq. \eqref{Eq:loss_upper_bound}, we can observe that the augmentation results depend on the training data. If we directly apply eq. \eqref{Eq:loss_upper_bound}, we in fact mainly augment the majority classes, which disobeys our goal. 

Hence, to tackle this issue, we propose to unify the class-conditional weights in \cite{cui2019class} with eq. \eqref{Eq:loss_upper_bound} for down-weighting the losses of majority samples. The class-conditional weights are defined as $\epsilon_c \approx (1-\beta)/(1-\beta^{n_c})$, where $n_c$ is the number of data in $c$-th class and $\beta$ is the hype-parameter with a recommended value $(N-1)/N$. To sum up, the optimal parameters $\Theta^{*}$ can be calculated with the weighted loss on the training set:
\begin{align}
    \label{eq:update_theta}
    \Theta^{*}(\boldsymbol{\Sigma}) = \arg \mathop{\min}_{\Theta} \sum_{i=1}^{N} \epsilon_{i} L_{\infty}(f(\boldsymbol{x_i};\Theta),y_i; \boldsymbol{\Sigma})
\end{align}
If we treat covariance matrices $\boldsymbol{\Sigma}$ as training hyperparameters, we actually can search their optimal value on the validation set as \cite{L2RW,MetaWeightNet,andrychowicz2016learning,ravi2016optimization,ren2018meta}. Specifically, consider a small validation set $D^v=\{\x_i^v,y_i^v\}_{i=1}^{N^v}$, where $N^v$ is the amount of total samples and $N^v \ll N$. The optimal class-wise covariance can be obtained by minimizing the following validation loss:
\begin{align}
    \label{eq:update_sigma}
    \boldsymbol{\Sigma}^{*}=\arg\mathop{\min}_{\boldsymbol{\Sigma}}\sum_{i=1}^{N^v}L_{ce}(f(\x_i^v; \Theta^{*}(\boldsymbol{\Sigma})), y_i^v),
\end{align}where $L_{ce}(\cdot,\cdot)$ is the cross-entropy (CE) loss function. Since the validation set is balanced, we adopt vanilla CE loss to calculate the loss on validation set. 

\textbf{Online approximation}. To obtain the optimal value of $\Theta$ and $\boldsymbol{\Sigma}$, we need to go through two nested loops, which can be cost-expensive. Hence, we exploit an online strategy to update $\Theta$ and $\boldsymbol{\Sigma}$ through one-step loops.
Given current step $t$, we can obtain current covariance matrices $\boldsymbol{\Sigma}^t$ according to \cite{ISDA}. Next, we update the parameters of classifier $\Theta$ with following objective:
\begin{align}
    \widetilde{\Theta}^{t+1}(\boldsymbol{\Sigma^t}) \leftarrow \Theta^{t} - \alpha \nabla_{\Theta}  \sum_{i=1}^{N} \epsilon_{i} L_{\infty}(f(\boldsymbol{x_i};\Theta^t),y_i; \boldsymbol{\Sigma}^t),
\end{align}where $\alpha$ is the step size for $\Theta$. After executing this step of backpropagation, we obtain the optimized parameters $\widetilde{\Theta}^{t+1}(\boldsymbol{\Sigma^t})$. Then we can update the class-wise covariance $\boldsymbol{\Sigma}$ using the gradient produced by eq. \eqref{eq:update_sigma}:
\begin{align}
    \boldsymbol{\Sigma}^{t+1} \leftarrow \boldsymbol{\Sigma}^{t} - \gamma \nabla_{\boldsymbol{\Sigma}} \sum_{i=1}^{N^v} L_{ce}(f(\boldsymbol{x}_i^v; \widetilde{\Theta}^{t+1}(\boldsymbol{\Sigma}^t)), y_i^v),
\end{align}where $\gamma$ it the step size for $\boldsymbol{\Sigma}$. With the learned class-wise $\boldsymbol{\Sigma}$, we can ameliorate the parameters $\Theta$ of classifier as:
\begin{align}
    \Theta^{t+1} \leftarrow \Theta^{t} - \alpha \nabla_{\Theta}  \sum_{i=1}^{N} \epsilon_{i} L_{\infty}(f(\boldsymbol{x_i};\Theta^{t}),y_i; \boldsymbol{\Sigma}^{t+1}).
\end{align}
Since the updated class-wise covariance matrices $\boldsymbol{\Sigma}^{t+1}$ are learned from balanced validation data, we could expect $\boldsymbol{\Sigma}^{t+1}$ help to learn better classifier parameters $\Theta^{t+1}$. In practice, we adopt the generally used technique SGD to implement our algorithm. In addition, several previous works have demonstrated that training the networks without re-balancing strategy in the early stage learns better generalizable representations \cite{kang2019decoupling,cao2019learning,jamal2020rethinking}. Hence, we first train classifiers with vanilla CE loss, then with MetaSAug. The training algorithm is shown in Algorithm. \ref{alg:1}. 

\subsection{Discussion} 
\begin{figure}[htbp]
\centering
\includegraphics[width=1\linewidth]{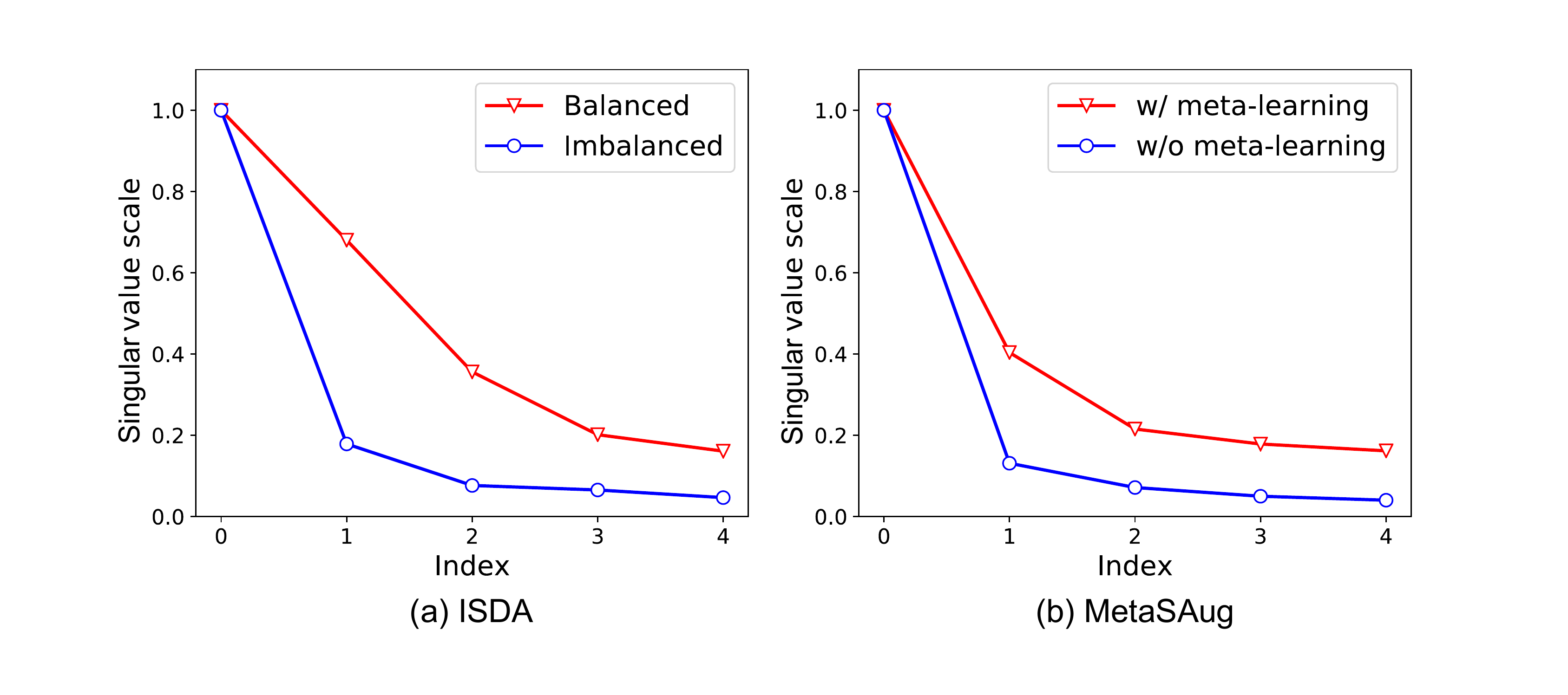}
\caption{The top-5 singular values (max normalized) of covariance matrix $\boldsymbol{\Sigma}_r$ learned by ISDA and MetaSAug. \textbf{(a)}: ``Balanced" refer to the covariance matrix estimated on balanced training set (CIFAR-10), while ``Imbalanced" implies the $\boldsymbol{\Sigma}_r$ estimated on imbalanced training set (CIFAR-LT-10 with imbalance factor=$200$). \textbf{(b)}: Both experiments are conducted on the imbalanced set. The red and blue lines denote the $\boldsymbol{\Sigma}_r$ learned by MetaSAug with and without our meta-learning method, respectively.}
\label{fig:eigen}
\vspace{-3mm}
\end{figure}

In this work, we utilize meta-learning to learn proper covariance matrices for augmenting the minority classes.
Hence, it's essential to find out what $\boldsymbol{\Sigma}$ have learned from the validation data. To investigate this question, we conduct singular value decomposition to extract the singular values for the covariance matrix $\boldsymbol{\Sigma}_r$ of the most rare class $r$:
\begin{align}
    \boldsymbol{\Sigma}_r = \boldsymbol{\mathrm{UMV}}^{\top},
\end{align}where each element in the diagonal of $\boldsymbol{\mathrm{M}}$ is the singular value of $\boldsymbol{\Sigma}_r$.
Then, we illustrate the top-5 singular values (max-normalized) of $\boldsymbol{\Sigma}_r$ learned by ISDA and our MetaSAug. Principal component analysis demonstrates that the eigenvector with larger singular value will contain more information variations \cite{wold1987principal,abdi2010principal}. From Fig. \ref{fig:eigen}(a), we observe that the largest singular value of $\boldsymbol{\Sigma}_r$ learned by ISDA on imbalanced dataset is remarkably larger than other singular values, while the information signals of other eigenvectors with smaller singular values are suppressed. 
This sharp distribution of singular values implies that the $\boldsymbol{\Sigma}_r$ has less important principal components, which one may not be able to sample diversified transformation vectors with the $\boldsymbol{\Sigma}_r$. The reason is that ISDA can not estimate appropriate $\boldsymbol{\Sigma}_r$ with the scarce data of minority classes. 

When MetaSAug applies the proposed meta-learning method to learn $\boldsymbol{\Sigma}_r$, the singular value distribution of $\boldsymbol{\Sigma}_r$ becomes relatively balanced, as shown in Fig. \ref{fig:eigen}(b). Apart from the one with largest singular value, the other eigenvectors also contain great information variance. With the $\boldsymbol{\Sigma}_r$ that has a balanced singular value distribution, one may sample diverse transformation vectors, leading to better augmentation results.
In summary, with our proposed meta-learning method, the learned covariance matrix $\boldsymbol{\Sigma}_r$ contains more important principal components, implying that it may contain more semantic directions. 
In addition, we further experimentally verify that our meta-learning method can help improve the performance of classifiers in Section \ref{ablation study}.

\begin{algorithm}[t]
\caption{Leaning algorithm of MetaSAug} 
\label{alg:1}
\hspace*{0.02in} {\bf Input:} 
Training set $D$; validation set $D^v$; ending steps $T_1$ and $T_2$;

\hspace*{0.02in} {\bf Output:} 
Learned classifier parameter $\Theta$
\begin{algorithmic}[1]
\For{t $\leq$ $T_1$} 
\State Sample a batch $B=\{(\boldsymbol{x}_i, y_i)\}_{i=1}^{|B|}$ from $D$
\State Calculate loss $\mathcal{L}_{B}=\frac{1}{|B|}\sum_{i=1}^{|B|}L_{ce}(f(\boldsymbol{x}_i;\Theta),y_i)$
\State Update $\Theta\leftarrow\Theta-\alpha\nabla_{\Theta}\mathcal{L}_{B}$ 
\EndFor
\For{$T_1$ $\textless$ t $\leq T_2$}
\State Sample a batch $B=\{(\boldsymbol{x}_i, y_i)\}_{i=1}^{|B|}$ from $D$
\State Sample a batch $B^v=\{(\boldsymbol{x}_i^v, y_i^v)\}_{i=1}^{|B^v|}$ from $D^v$
\State Obtain current covariance matrices $\boldsymbol{\Sigma}$
\State Compute $\mathcal{L}_{B} =  \sum_{i=1}^{|B|} \epsilon_{i} L_{\infty}(f(\boldsymbol{x}_i;\Theta),y_i; \boldsymbol{\Sigma})$
\State Update $\widetilde{\Theta}(\boldsymbol{\Sigma}) \leftarrow \Theta - \alpha \nabla_{\Theta} \mathcal{L}_{B}$
\State Compute $\mathcal{L}_{B^v}= \sum_{i=1}^{|B^v|} L_{ce}(f(\boldsymbol{x}_i^v; \widetilde{\Theta}(\boldsymbol{\Sigma})), y_i^v)$
\State Update $\boldsymbol{\Sigma} \leftarrow \boldsymbol{\Sigma} - \gamma \nabla_{\boldsymbol{\Sigma}}\mathcal{L}_{B^v} $
\State Calculate the loss with the updated $\boldsymbol{\Sigma}$ 
    
    $\widetilde{\mathcal{L}}_{B}=\sum_{i=1}^{|B|} \epsilon_{i} L_{\infty}(f(\boldsymbol{x}_i;\Theta),y_i; \boldsymbol{\Sigma})$
\State update $\Theta \leftarrow \Theta - \alpha \nabla_{\Theta}\widetilde{\mathcal{L}}_{B}$
\EndFor
\end{algorithmic}
\end{algorithm}

\section{Experiment}
We evaluate our method on the following long-tailed datasets: CIFAR-LT-10, CIFAR-LT-100, ImageNet-LT, iNaturalist 2017 and iNaturalist 2018. In addition, we report the average result of 3 random experiments. For those experiments conducted in the same settings, we directly quote their results from original papers. Code is available at \url{https://github.com/BIT-DA/MetaSAug}.

\subsection{Datasets}
\textbf{Long-Tailed CIFAR} is the long-tailed version of CIFAR dataset. The original CIFAR-10 (CIFAR-100) dataset consists of 50000 images drawn from 10 (100) classes with even data distribution. In other words, CIFAR-10 (CIFAR-100) has 5000 (500) images per class. Following \cite{cui2019class}, we discard some training samples to construct imbalanced datasets. We build 5 training sets by varying imbalance factor $\mu \in \{200, 100, 50, 20, 10\}$, where the $\mu$ denotes the image amount ratio between largest and smallest classes. If let $n_i$ denotes the sample amount of $i$-th class, we can define $\mu = \frac{\max_{i}(n_i)}{\min_{j}(n_j)}$. As for test sets, we use the original balanced test sets. Following \cite{jamal2020rethinking}, we randomly select ten samples per class from training set to construct validation set $D^v$.

\textbf{ImageNet-LT}: ImageNet \cite{imagenet2014} is a classic visual recognition dataset, which contains 1,281,167 training images and 50,000 validation images. Liu et al. \cite{liu2019large} build the long-tailed version of ImageNet, termed as ImageNet-LT. After discarding some training examples, ImageNet-LT remains 115,846 training examples in 1,000 classes. The imbalance factor is 1280/5. We adopt the original validation to test methods. In addition, Liu et al. \cite{liu2019large} also construct a small balanced validation set with 20 images per class. Hence, we adopt ten images per class from it to construct our validation set $D^v$ as \cite{jamal2020rethinking}.

\textbf{iNaturalist 2017 and iNaturalist 2018}. The iNaturalist datasets are large-scale datasets with images collected from real-world, which have an extremely imbalanced class distribution. The iNatualist 2017 \cite{inaturalist2017} includes 579,184 training images in 5,089 classes with an imbalance factor of 3919/9, while the iNaturalist 2018 \cite{inaturalist} is composed of 435,713 images from 8,142 classes with an imbalance factor of 1000/2. We adopt the original validation set to test our method. To construct the validation set $D^v$, we select five and two images from the training sets of iNaturalist 2017 and iNaturalist 2018, respectively, following \cite{jamal2020rethinking}.

\subsection{Visual Recognition on CIFAR-LT}
We conduct comparison experiments on the long-tailed datasets CIFAR-LT-10 and CIFAR-LT-100. Following \cite{jamal2020rethinking, cui2019class}, we adopt the ResNet-32 \cite{resnet} as the backbone network in our experiments. 

\textbf{Implementation details}.
For the baselines LDAM and LDAM-DRW, we reproduce them with the source code released by authors \cite{cao2019learning}. We train the ResNet-32 \cite{resnet} with standard stochastic gradient decent (SGD) with momentum $0.9$ and weight decay of $5 \times 10^{-4}$ for all experiments. And We train the models on a single GPU for 200 epochs. In addition, we decay the learning rate by 0.01 at the 160$^{th}$ and 180$^{th}$ epochs as \cite{jamal2020rethinking}. For our method, we adopt the initial learning rate 0.1. And we set the batch size as 100 for our experiments. The hyperparameter $\lambda$ is selected from $\{0.25, 0.5, 0.75, 1.0\}$.

\textbf{Comparison methods}. We compare our method with the following methods:
\begin{itemize}
	\item \textbf{Cross-entropy training} is the baseline method in long-tailed visual recognition, which trains ResNet-32 using vanilla cross-entropy loss function.
	\item \textbf{Class weighting methods}. This type of method assigns weights to training examples in class level, which includes class-balanced loss \cite{cui2019class}, meta-class-weight \cite{jamal2020rethinking} and LDAM-DRW \cite{cao2019learning}. Class-balanced loss proposes effective number to measure the sample size of each class and the class-wise weights. Class-balanced focal loss and class-balanced cross-entropy loss refer to applying class-balanced loss on focal loss and cross-entropy loss, respectively. LDAM-DRW allocates label-aware margins to the examples based on the label distribution, and adopts deferred re-weighting strategy for better performance on tail classes.
	\item \textbf{Instance weighting methods} assign weights to samples according to the instance characteristic \cite{L2RW,MetaWeightNet,lin2017focal}. For example, focal loss \cite{lin2017focal} determine the weights for samples based on the sample difficulty. Though focal loss is not specially designed for long-tailed classification, it can penalize the samples of minority classes if the classifier overlooks the minority classes during training.  

	\item \textbf{Meta-learning methods}. In fact, these methods adopt meta-learning to learn better class level or instance level weights \cite{jamal2020rethinking,MetaWeightNet,L2RW}. For saving space, we only introduce them here. Meta-class-weight \cite{jamal2020rethinking} exploits meta-learning to model the condition distribution difference between training and testing data, leading to better class level weights. While L2RW \cite{L2RW} and meta-weight-net \cite{MetaWeightNet} adopt meta-learning to model the instance-wise weights. L2RW directly optimizes the weight variables, while meta-weight-net additionally constructs a multilayer perceptron network to model the weighting function. Note that both L2RW and meta-weight-net can handle the learning with imbalanced label distribution and noisy labels.
	\item \textbf{Two-stage methods}. We also compared with methods that adopt two-stage learning \cite{jamal2020rethinking,cao2019learning,cui2018large}. BBN \cite{BBN} unifies the representation and classifier learning stages to form a cumulative learning strategy.
\end{itemize}

\begin{table*}
	\centering
	\caption { Test top-1 errors (\%) of ResNet-32 on CIFAR-LT-10 under different imbalance settings. * indicates results reported in original paper. $\dag$ indicates results reported in \cite{jamal2020rethinking}.}
	\vspace{2pt}{%
		\begin{tabular}{l|c|c|c|c|c}
			\hline
			Imbalance factor & 200 & 100 & 50 & 20 & 10\\ \hline
			Cross-entropy training & 34.13 & 29.86  & 25.06  &  17.56 & 13.82 \\ \hline
			Class-balanced cross-entropy loss \cite{cui2019class} & 31.23 & 27.32  & 21.87 & 15.44 & 13.10  \\ \hline
			Class-balanced fine-tuning$^{\dag}$ \cite{cui2018large} & 33.76 & 28.66 & 22.56 & 16.78 & 16.83  \\ \hline
			BBN$^{*}$ \cite{BBN} & - & 20.18 & 17.82 & - & 11.68 \\ \hline
			Mixup \cite{mixup} (results from \cite{BBN}) & - & 26.94 & 22.18 & - & 12.90 \\ \hline
			L2RW$^{\dag}$ \cite{L2RW} & 33.75 & 27.77 & 23.55 & 18.65 & 17.88 \\ \hline
			Meta-weight net$^{\dag}$ \cite{MetaWeightNet} & 32.80 & 26.43 & 20.90 & 15.55 & 12.45  \\ \hline
			Meta-class-weight with cross-entropy loss$^{\dag}$ \cite{jamal2020rethinking} & 29.34 & 23.59 & 19.49 & 13.54 & 11.15 \\\hline
			\textbf{MetaSAug with cross-entropy loss} & \textbf{23.11} & \textbf{19.46} & \textbf{15.97} & \textbf{12.36} & \textbf{10.56} \\ \hline\hline 
			Focal loss$^{\dag}$ \cite{lin2017focal} & 34.71 & 29.62 & 23.29 & 17.24 & 13.34  \\ \hline
			Class-balanced focal loss$^{\dag}$ \cite{cui2019class} & 31.85 & 25.43 & 20.78 & 16.22 & 12.52 \\ \hline
			Meta-class-weight with focal loss$^{\dag}$ \cite{jamal2020rethinking} & 25.57 & 21.10 & 17.12 & 13.90 & 11.63 \\ \hline
			\textbf{MetaSAug with focal loss} & \textbf{22.73} & \textbf{19.36} & \textbf{15.96} & \textbf{12.84} & \textbf{10.74} \\ \hline\hline 
			LDAM loss\cite{cao2019learning} & 33.25 & 26.45 & 21.17 & 16.11 & 12.68  \\ \hline
			LDAM-DRW \cite{cao2019learning} & 25.26 & 21.88 & 18.73 & 15.10 & 11.63  \\ \hline
			Meta-class-weight with LDAM loss $^{\dag}$ \cite{jamal2020rethinking} & 22.77 & 20.00 & 17.77 & 15.63 & 12.60 \\ \hline
			
			\textbf{MetaSAug with LDAM loss}  & \textbf{22.65} & \textbf{19.34} & \textbf{15.66} & \textbf{11.90} & \textbf{10.32} \\ \hline
		\end{tabular}
	}%
	\label{tab:CIFAR-10}
	
\end{table*}

\begin{table*}
	\centering
	\caption {Test top-1 errors (\%) of ResNet-32 on CIFAR-LT-100 under different imbalance settings. * indicates results reported in original paper. $\dag$ indicates results reported in \cite{jamal2020rethinking}.}
	{%
		\begin{tabular}{l|c|c|c|c|c}
			\hline
			Imbalance factor & 200 & 100 & 50 & 20 & 10 \\ \hline
			
			Cross-entropy training & 65.30 & 61.54 & 55.98 & 48.94 & 44.27 \\ \hline
			Class-balanced cross-entropy loss \cite{cui2019class} & 64.44 & 61.23 & 55.21 & 48.06 & 42.43 \\ \hline
			Class-balanced fine-tuning$^{\dag}$ \cite{cui2018large} & 61.34 & 58.50 & 53.78 & 47.70 & 42.43 \\ \hline
			BBN$^*$ \cite{BBN} & - & 57.44 & 52.98 & - & 40.88 \\ \hline
			Mixup \cite{mixup} (results from \cite{BBN}) & - & 60.46 & 55.01 & - & 41.98 \\ \hline
			L2RW$^{\dag}$ \cite{L2RW} & 67.00 & 61.10 & 56.83 & 49.25 & 47.88 \\ \hline
			Meta-weight net$^\dag$ \cite{MetaWeightNet} & 63.38 & 58.39 & 54.34 & 46.96 & 41.09 \\ \hline
			Meta-class-weight with cross-entropy loss$^\dag$ \cite{jamal2020rethinking} & 60.69 & 56.65 & 51.47 & 44.38 & 40.42 \\ \hline
			\textbf{MetaSAug with cross-entropy loss} & \textbf{60.06} & \textbf{53.13} & \textbf{48.10} & \textbf{42.15} & \textbf{38.27} \\ \hline\hline 
		
			Focal loss$^\dag$ \cite{lin2017focal} & 64.38 & 61.59 & 55.68 & 48.05 & 44.22  \\ \hline
		
			Class-balanced focal loss$^\dag$ \cite{cui2019class} & 63.77 & 60.40 & 54.79 & 47.41 & 42.01  \\ \hline
			Meta-class-weight with focal loss$^\dag$ \cite{jamal2020rethinking}& 60.66 & 55.30 & 49.92 & 44.27 & 40.41 \\ \hline
			\textbf{MetaSAug with focal loss} & \textbf{59.78} & \textbf{54.11} & \textbf{48.38} & \textbf{42.41} & \textbf{38.94} \\ \hline\hline 
		
			LDAM loss \cite{cao2019learning}  & 63.47 & 59.40  & 53.84 & 48.41 & 42.71  \\ \hline
			LDAM-DRW \cite{cao2019learning} & 61.55 & 57.11 & 52.03 & 47.01 & 41.22 \\ \hline
			Meta-class-weight with LDAM loss$^\dag$ \cite{jamal2020rethinking} & 60.47 & 55.92 & 50.84 & 47.62 & 42.00 \\ \hline
			\textbf{MetaSAug with LDAM loss} & \textbf{56.91} & \textbf{51.99} & \textbf{47.73} & \textbf{42.47} & \textbf{38.72} \\ \hline
		\end{tabular}
	}
	\label{tab:CIFAR-100}
\end{table*}

\textbf{Results}. 
\label{comparision_results}
The experimental results of long-tailed CIFAR-10 with different imbalance factors are shown in Table \ref{tab:CIFAR-10}, which are organized into three groups according to the adopted basic losses (i.e., cross-entropy, focal, and LDAM). 

From the results, we can observe that re-weighting strategies are effective for the long-tailed problems, since several re-weighting methods (e.g., meta-class-weight) outperform the cross-entropy training by a large margin. We evaluate our method with the three basic losses. The results reveal that our method can consistently improve the performance of the basic losses significantly. Particularly, MetaSAug notably surpasses mixup that conducts augmentation on the inputs, manifesting that facilitating semantic augmentation is more effective in long-tailed scenarios. In addition, our method performs better than the re-weighting methods. This demonstrates that our augmentation strategy indeed can ameliorate the performance of classifiers. When the dataset is less imbalanced (implying imbalance factor=10), our method can still stably achieve performance gains, revealing that MetaSAug won't 
damage the performance of classifier under the relatively balanced setting.

Table \ref{tab:CIFAR-100} presents the classification error of dataset long-tailed CIFAR-100, from which we can still observe that our methods achieve the best results in each group. Particularly, ``MetaSAug with LDAM loss" exceeds the best competing method ``Meta-class-weight with LDAM loss" by 3.56\%.

\begin{figure*}[htbp]\centering
	\setlength{\abovecaptionskip}{0.cm}
	\setlength{\belowcaptionskip}{0.cm}
	\centering
	\includegraphics[width=1.0\textwidth]{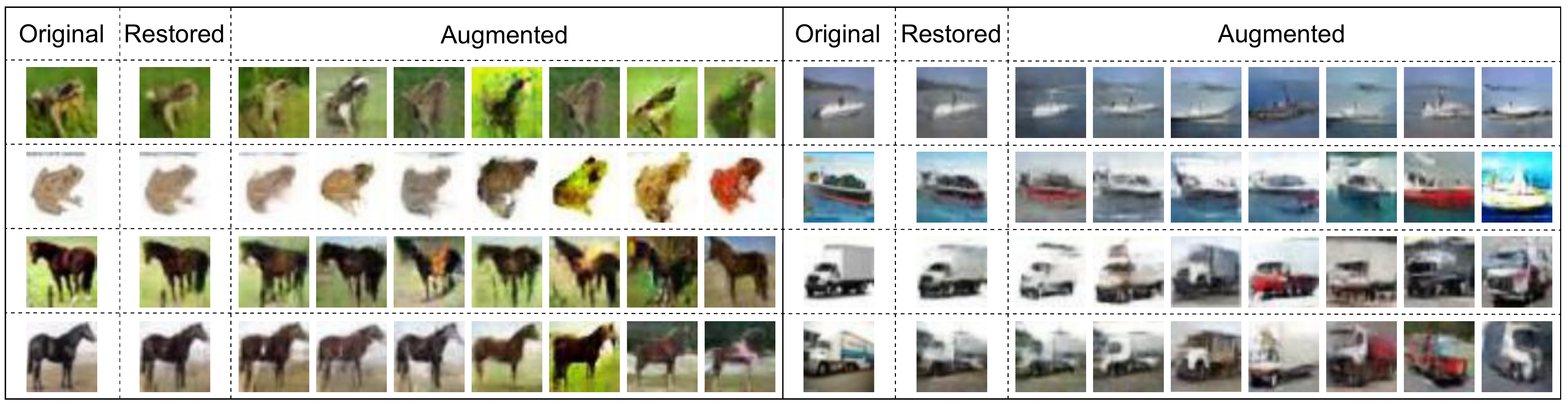}
	\caption{Visualization of the augmented examples for the four rarest classes: frog, horse, ship and truck (frequent $\rightarrow$ rare). We adopt WGAN-GP \cite{WGAN-GP} generator to search the images corresponding to the augmented features. ``Original'' refers to the original training samples. ``Restored'' and ``Augmented" present the original and augmented images generated by the generator, respectively. Our method is able to semantically alter the semantic of training images, e.g., changing color of objects, backgrounds, shapes of objects, etc.}
	\label{fig:visualization}
	
\end{figure*}

\begin{figure*}[htbp]\centering
	\setlength{\abovecaptionskip}{0.cm}
	\setlength{\belowcaptionskip}{0.cm}
	\centering
	\includegraphics[width=1.0\textwidth]{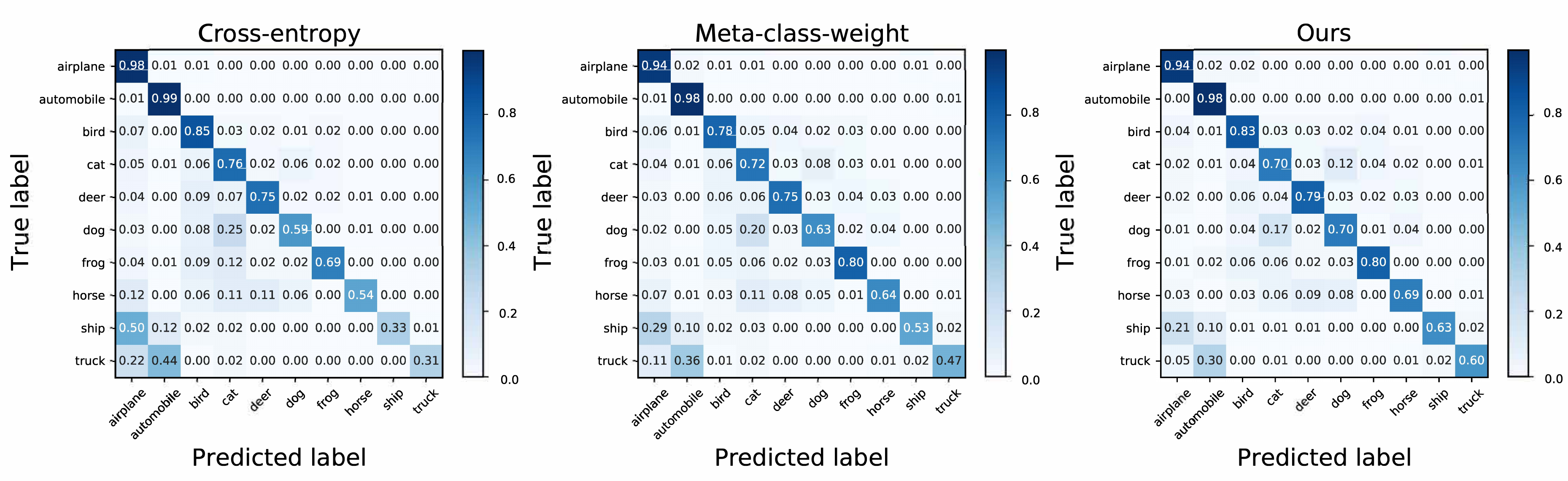}
	\caption{Illustration of confusion matrices of the vanilla cross-entropy training, meta-class-weight \cite{jamal2020rethinking}, and our method on dataset CIFAR-LT-10. The imbalance factor is 200. Classes are ranked by the frequency, i.e., frequent (left) $\rightarrow$ rare (right).}
	\label{Fig_confusion_matrix}
\end{figure*}

\subsection{Visual Recognition on ImageNet-LT} \vspace{-1mm}
We use ResNet-50 \cite{resnet} as the backbone network in the experiments on ImageNet-LT. And we train ResNet-50 with batch size 64. We decay the learning rate by 0.1 at 60$^{th}$ epoch and 80$^{th}$ epoch. In addition, for training efficiency, we only finetune the last full-connected layer while fixing the representations in the meta-learning stage. We reproduce the comparison methods based on the code released by authors.

\textbf{Results}. The experimental results are shown in Table \ref{tab:ImageNet-LT}. Class-balanced cross-entropy performs better than cross-entropy training and LDAM-DRW surpasses LDAM by a large margin. These results imply that re-weighting strategy is also effective for the dataset with a large number of classes. Hence, MetaSAug also adopts this strategy to better fulfill the semantic augmentation procedure. In addition, compared with the best competing method meta-class-weight, MetaSAug can still yield better results, demonstrating that MetaSAug is able to perform data augmentation useful for the classification learning of classifiers.

\begin{table}
	\centering
	\caption {Test top-1 error rate (\%) on ImageNet-LT of different models. (CE=Cross-entropy)}
	{%
		\begin{tabular}{l|c}
			\hline
			Method & Top-1 error \\ \hline
			CE training & 61.12 \\ \hline
			Class-balanced CE \cite{cui2019class}  & 59.15 \\ \hline
			OLTR \cite{liu2019large} & 59.64 \\ \hline
			LDAM \cite{cao2019learning} & 58.14 \\ \hline
			LDAM-DRW \cite{cao2019learning} & 54.26 \\ \hline
			Meta-class-weight with CE loss \cite{jamal2020rethinking} & 55.08 \\ \hline
			\textbf{MetaSAug with CE loss} & \textbf{52.61} \\ \hline
		\end{tabular}
	}%
	\label{tab:ImageNet-LT}
	
\end{table}

\subsection{Visual Recognition on iNaturalist Datasets}
For fair comparisons, we adopt ResNet-50 \cite{resnet} as the backbone network for iNaturalist 2017 and 2018. Following \cite{jamal2020rethinking}, we pre-train the backbone network on ImageNet for iNaturalist 2017. As for iNaturalist 2018, the network is pre-trained on ImageNet and iNaturalist 2017. We use stochastic gradient descent (SGD) with momentum to train models. The batch size is set as 64 and the initial learning rate is 0.01. In the meta-learning stage of our method, we decay the learning rate to 0.0001 and only finetune the last fully-connected layer for training efficiency.

\textbf{Results}. Table \ref{tab:iNat} presents the experimental results on the naturally-skewed datasets iNaturalist 2017 and iNaturalist 2018. MetaSAug and meta-class-weight both exploit the CE loss as basic loss. Compared with the improvement brought by class-balanced CE \cite{cui2019class} to CE Loss, MetaSAug further enhances the performance of CE loss, implying that performing effective data augmentation is also of importance for long-tailed classification. In addition, MetaSAug yields the best results among these competitive methods on iNat 2018 and is on par with the state-of-art method BBN \cite{BBN} on iNat 2017. These results demonstrate that our method indeed can facilitate data augmentation useful for classification in the deep feature space.

\begin{table}
	\centering
	\caption {Test top-1 error rate (\%) on iNaturalist (iNat) 2017 and 2018 of different models. $^{*}$results is quoted from original papers. $\dag$ indicates results reported in \cite{jamal2020rethinking}. (CE=Cross-entropy)}
	{%
		\begin{tabular}{l|c|c}
			\hline
			Method & iNat 2017 & iNat 2018 \\ \hline
			CE & 43.21 & 34.24 \\ \hline
			Class-balanced CE \cite{cui2019class} & 42.02 & 33.57 \\ \hline
			Class-balanced focal$^*$ \cite{cui2019class} & 41.92 & 38.88 \\ \hline
			cRT$^*$ \cite{kang2019decoupling} & - & 32.40 \\ \hline
			BBN$^*$ \cite{BBN} & \textbf{36.61} & 33.71 \\ \hline
			LDAM$^*$ \cite{cao2019learning}  & - & 35.42 \\ \hline
			LDAM \cite{cao2019learning} & 39.15 & 34.13 \\ \hline
			LDAM-DRW$^*$ \cite{cao2019learning} & - & 32.00  \\ \hline
			LDAM-DRW \cite{cao2019learning} & 37.84 & 32.12 \\ \hline
			Meta-class-weight$^{\dag}$ \cite{jamal2020rethinking} & 40.62 & 32.45 \\ \hline
			\textbf{MetaSAug} & 36.72 & \textbf{31.25}\\ \hline
		\end{tabular}
	}%
	\label{tab:iNat}
\end{table}

\subsection{Analysis}
\label{ablation study}
\textbf{Ablation study}. To verify each component of MetaSAug, we conduct ablation study (see Table \ref{tab:ablation}). Removing re-weighting or meta-learning causes performance drop. This manifests 1) re-weighting is effective to construct a proper meta-learning objective, and 2) our meta-learning method can indeed learn covariance useful for classification. Importantly, the latter is non-trivial since MetaSAug achieves the notable accuracy gains as shown in Table \ref{tab:ablation}. While this cannot be reached by meta-weighting methods with fixed ISDA (e.g., L2RW; Meta-weight net, MWN; Meta-class-weight, MCW). Furthermore, we observe that ISDA can boost former long-tailed methods to some extent, but the improvement is limited. This also validate the importance of our meta-learning algorithm.

\textbf{Adaptivity to deeper backbone networks}. For a reasonable comparison with baselines, we adopt the commonly used ResNet-32 and ResNet-50 to evaluate our method on CIFAR-LT and ImageNet-LT/iNaturalist datasets, respectively. However, MetaSAug can be easily adapted to other networks, and, as indicated in \cite{ISDA}, deeper models may even benefit our method more due to their stronger ability to model complex semantic relationships. In Table \ref{tab:deeper_backbones}, we show the results of MetaSAug, MCW \cite{jamal2020rethinking} and LDAM-DRW \cite{cao2019learning} with different backbone networks. One can observe that MetaSAug consistently outperforms other methods.

\textbf{Confusion matrices}. To find out whether our method ameliorates the performance on minority classes, we plot the confusion matrices of cross-entropy (CE) training, meta-class-weight \cite{jamal2020rethinking}, and our method on CIFAR-LT-10 with imbalance factor 200. From Fig.\ref{Fig_confusion_matrix}, we can observe that CE training can almost perfectly classify the samples in majority classes, but suffers severe performance degeneration on the minority classes. Due to the proposed two-component weighting, meta-classes-weight performs much better than CE training on the minority classes. Since MetaSAug inclines to augment the minority classes, it can further enhance the performance on rare classes and reduce the confusion between similar classes (implying automobile \& truck, and airplane \& ship).

\textbf{Visualization Results}. To intuitively reveal that our method can indeed alter the semantics of training examples and generate diverse meaningful augmented samples, we carry out the visualization experiment (the detailed visualization algorithm is presented in \cite{ISDA}). The visualization results are shown in Fig. \ref{fig:visualization}, from which we can observe that MetaSAug is capable of semantically altering the semantics of training examples while preserving the label identity. Particularly, MetaSAug can still generate meaningful augmented samples for the rarest class ``truck".

\begin{table}
	\centering
	\caption {Ablation study of MetaSAug using cross-entropy loss on dataset CIFAR-LT-10. The results are top-1 errors (\%).}
	{%
		\begin{tabular}{l|c|c|c}
			\hline
			Imbalance factor  & 100 & 50 & 20 \\ \hline
			MetaSAug w/o re-weighting & 25.96 & 20.63 & 15.15 \\ \hline
			MetaSAug w/o meta-learning & 21.68 & 17.43 & 13.08 \\ \hline
			MetaSAug & \textbf{19.46} & \textbf{15.97} & \textbf{12.36} \\ \hline  \hline
			ISDA, L2RW \cite{L2RW} & 25.16 & 20.78 & 16.53  \\ \hline
			ISDA, MWN \cite{MetaWeightNet} & 24.69 & 20.42 & 14.71 \\ \hline
			ISDA, MCW \cite{jamal2020rethinking} & 20.78 & 17.12 & 12.93 \\ \hline
		\end{tabular}
	}%
	\label{tab:ablation}
\end{table}

\begin{table}[htbp]
	\centering
	\small
	
	\caption {\small Test top-1 errors (\%) on ImageNet-LT of methods with different backbone networks.}
	
	{%
		\begin{tabular}{l|c|c|c}
			\hline
			Network & MCW \cite{jamal2020rethinking} & LDAM-DRW \cite{cao2019learning} &  MetaSAug \\ \hline
			ResNet-50 & 55.08 & 54.26 & \textbf{52.61}  \\ \hline
			ResNet-101 & 53.76 & 53.55  & \textbf{50.95}  \\ \hline
			ResNet-152 & 53.18 &  52.86 & \textbf{49.97}  \\ \hline
		\end{tabular}
		
	}%
	\vspace{-3mm}
	\label{tab:deeper_backbones} 
\end{table}

\section{Conclusion}
In this paper, we delve into the long-tailed visual recognition problem and propose to tackle it from a data augmentation perspective, which has not been fully explored yet. We present a meta semantic augmentation (MetaSAug) approach that learn appropriate class-wise covariance matrices for augmenting the minority classes, ameliorating the learning of classifiers. In addition, MetaSAug is orthogonal to several former long-tailed methods, e.g., LDAM and focal loss. Extensive experiments on several benchmarks validate the effectiveness and versatility of MetaSAug.

{\small
\bibliographystyle{ieee_fullname}
\bibliography{Reference_CVPR2021}
}

\end{document}